\documentclass{article}
\usepackage{arxiv}

\usepackage[utf8]{inputenc} 
\usepackage[T1]{fontenc}    
\usepackage{hyperref}       
\usepackage{url}            
\usepackage{booktabs}       
\usepackage{amsfonts}       
\usepackage{nicefrac}       
\usepackage{microtype}      
\usepackage{lipsum}
\usepackage{amsbsy}

\usepackage{subcaption}
\usepackage{amsthm}
\usepackage{amsmath}
\usepackage{wrapfig}
\usepackage{array,multirow,graphicx}
\newtheorem{definition}{Definition}[section]
\newtheorem{theorem}{Theorem}[section]
\newtheorem{lemma}{Lemma}[section]
\newtheorem{corollary}{Corollary}[section]
\newtheorem{example}{Example}[section]

\title{A fast topological approach for predicting anomalies in time-varying graphs}

\author{
    Umar Islambekov\\
    Department of Mathematics and Statistics\\
    Bowling Green State University\\
    \texttt{iumar@bgsu.edu} \\
        \And
    Hasani Pathirana\\
    Department of Mathematics and Statistics\\
    Bowling Green State University\\
    \texttt{pathirh@bgsu.edu} \\
   \And
  Omid Khormali \\
  Department of Mathematics\\
  University of Evansville\\
  \texttt{ok16@evansville.edu} \\  
    \And  
  C\"{u}neyt G\"{u}rcan Ak\c{c}ora \\
  Department of Computer Science\\
  University of Manitoba\\
  \texttt{cuneyt.akcora@umanitoba.ca} \\
   \And  
  Ekaterina Smirnova\\
 Department of Biostatistics\\
  Virginia Commonwealth University\\
  \texttt{ekaterina.smirnova@vcuhealth.org} \\
}

\begin{document}
\maketitle

\vspace{-0.5cm}

\begin{abstract}
Large time-varying graphs are increasingly common in financial, social and biological settings. Feature extraction that
efficiently encodes the complex structure of sparse, multi-layered, dynamic graphs presents computational and methodological
challenges. In the past decade, a persistence diagram (PD) from topological data analysis (TDA) has become a popular descriptor of
shape of data with a well-defined distance between points. However, applications of TDA to graphs, where there is no intrinsic concept
of distance between the nodes, remain largely unexplored. This paper addresses this gap in the literature by introducing a
computationally efficient framework to extract shape information from graph data. Our framework has two main steps: first, we compute
a PD using the so-called lower-star filtration which utilizes quantitative node attributes, and then vectorize it by averaging the
associated Betti function over successive scale values on a one-dimensional grid. Our approach avoids embedding a graph into a
metric space and has stability properties against input noise. In simulation studies, we show that the proposed vector summary leads
to improved change point detection rate in time-varying graphs. In a real data application, our approach provides up to 22\% gain in
anomalous price prediction for the Ethereum cryptocurrency transaction networks.
\end{abstract}


\keywords{Topological data analysis \and Persistent homology \and Persistence diagram \and Betti function \and Anomaly detection}


\section{Introduction}
\label{sec:introduction}

Blockchain is an emerging technology that enabled multiple digital management applications, including the Bitcoin cryptocurrency systems. Blockchain uses an underlying peer-to-peer network to transmit blocks and proposed transactions between blockchain users worldwide, which allows to construct a global network transaction graph and relate its properties to price dynamics. The increasing worldwide popularity of the blockchain system coupled with highly informative and novel publicly available data presents novel opportunities for studying the current state of the global economy. 

Ethereum is the most popular blockchain platform, allowing smart contracts and enabling everyone to create a crypto-asset. However, studying the Ethereum network and utilizing its data presents computational and methodological challenges. Indeed, the Ethereum transaction graph is multi-layered, sparse, and dynamic. Nodes (i.e., account addresses) appear and disappear (i.e., no future transaction) daily, while the number of transactions widely fluctuates across days. Furthermore, the graph is large; even a daily snapshot of the Ethereum graph contains more than one million transactions. Analysis of blockchain data requires novel approaches in complex network mining with a task-specific focus, such as  community detection~\cite{victor2020address} for clustering similar blockchain investors, core decomposition~\cite{victor2021alphacore} for detecting influential users, and centrality algorithms~\cite{pontiveros2019mint} for understanding the flow of coins. Traditional approaches scale poorly and exhibit poor performance due to the disconnected, sparse and dynamic nature of the Ethereum graph. In financial research, blockchains have become an active research area with applications in asset price prediction and anomaly detection~\cite{abay2019chainnet,li2020dissecting}.  

Anomalous change in asset price is a major indicator of global economic changes, making the prediction of Ethereum price anomaly detection a problem of significant interest. An Ethereum network graph contains crucial information about daily asset transactions that can be used for anomalous price prediction. However, a graph, which is a data structure consisting of nodes and edges (connections between nodes), does not easily lend itself as a suitable input object to most predictive algorithms. Therefore, as the first step toward analyzing graph data, one needs to extract a set of informative quantitative features, such as a finite-dimensional vector, that is more tractable for applications. If a graph has node or edge attributes, the extracted features may also incorporate this information. In principle, such features should not only adequately encode the graph structure but also be robust to input noise and computationally efficient. 

In the present work, we develop a framework to study graph data using the tools of topological data analysis (TDA), which has become a popular approach in recent years to study the shape structure of data \cite{edelsbrunner2010computational,carlsson2009topology}. The main purpose of TDA is to quantify the underlying shape information by systematically analyzing features within data such as connected components, holes, and voids at various scale resolutions \cite{zomorodian2005computing,edelsbrunner2008persistent}. While it is more natural and common to apply TDA to a point cloud (a collection of points with a notion of distance between them) or an image, in theory, the scope of TDA is much wider. It encompasses other complex data types such as graphs where there is no intrinsic concept of distance between the nodes. However, despite many successful applications of TDA in variety of areas such as image analysis \cite{qaiser2019fast, smith2021topological}, genomics \cite{carriere2020topological}, time series  \cite{umeda2017time}, material science \cite{pike2020topological}, transportation \cite{li2019topological} and chemistry \cite{smith2021topological}, 
its utility for the data represented as graphs still remains largely under-explored. 

In analyzing graph data, the TDA-based framework has four notable aspects that jointly lack in most standard methods for graph analysis. First, TDA not only employs the information about nodes and edges but also considers higher-order structures such as triangles (formed by three edges) and tetrahedrons (formed by four triangles). 
Second, TDA can use quantitative node attributes to extract topological information from a graph. 
These node attributes may be supplied together with data (e.g., the amounts sent from each node representing an account address) or computed from a graph (e.g., the degree or betweenness centrality of each node). 
Third, TDA allows us to study a graph at multiple resolutions by tracking the evolution of its subgraphs over a range of scale values. Finally, due to the multi-scale perspective, the TDA features are known to be robust to input noise \cite{cohen2007stability,chazal2014persistence,cohen2010lipschitz,bubenik2015statistical,adams2017persistence}. In applications, the extracted topological features can be used as the only inputs or combined with other standard graph features. In practice, the latter use of TDA is common where it complements the information about data captured by other methods that lack a topological perspective \cite{chazal2021introduction}. While it is possible to embed a graph into a metric space by introducing a concept of distance between its nodes \cite{hajij2017persistent} and then apply TDA to the resulting point cloud, in our work, we instead apply TDA directly to a graph which has node attributes. Our approach does not lose the edge information and generally has a lower computational cost to implement.


The standard multi-scale summary of the shape of data computed via the TDA procedure is called a \textit{persistence diagram} (PD). Subsequently, persistence diagrams are often inputted as predictive features within a learning task~\cite{chazal2021introduction}. However, the non-vector form of persistence diagrams greatly limits their utility in practice. A typical remedy is to further extract a functional summary, such as a persistence landscape~\cite{bubenik2015statistical}, a persistence image \cite{adams2017persistence} and a Betti function\footnote{also called a \emph{Betti curve}. We will adopt the former for the rest of the paper.} \cite{chazal2021introduction,chung2022persistence} from a persistence diagram and then vectorize it to obtain a finite-dimensional vector. A classic way to vectorize a Betti function is to form a vector of its values taken at multiple evenly spaced scale locations. However, this method misses the behavior of a Betti function between two adjacent scale values. To preserve the functional form of a Betti function in the resulting vector summary, one would need to construct a dense grid of scale locations, which tends to be computationally costly and thus less practical in applications dealing with large datasets. To address these challenges, we introduce a novel and computationally efficient approach to vectorize a Betti function by averaging its values between the two scales using integration (see Section~\ref{subsec:VAB} for more details).   

Our new vectorization scheme can be used in two different settings. If the goal is to preserve the information about the functional form of a Betti function in the resulting vector summary, our method, similar to the traditional way, can capture essentially the same topological information. In this case, the extracted vector summary will be high-dimensional and costly to compute. However, unlike the classic approach, our method is still applicable if one needs to extract an informative low-dimensional vector representation using a sparse grid of scale values.

Through simulation studies, we demonstrate the utility of our vector summary to analyze data represented as graphs. In the first simulation, we compare the two ways of constructing Betti-function-based vector summaries in the context of the change point detection problem for time-evolving graphs. In the second study, we integrate our vector summary with a statistical concept of data depth to find anomalous time points in cryptocurrency transaction graphs. 

The rest of the paper is organized as follows. 
Section~\ref{sec:theory} provides a brief overview of TDA, the proposed framework to extract topological vector summaries from graphs and the associated stability results. In Section~\ref{sec:experiments}, we present the results of two experimental studies. We conclude the paper with Section~\ref{sec:conclusion} summarizing the contributions and proposing directions for future research. We open-source our {\tt R} code at \url{https://github.com/TDAvecs/VAB}. 

\section{Topological Data Analysis on Graphs}
\label{sec:theory}

\subsection{Background information on TDA}
\begin{wrapfigure}{r}{0.45\textwidth}
\begin{center}
    \vspace{-0.75cm}
    \includegraphics[scale=0.3]{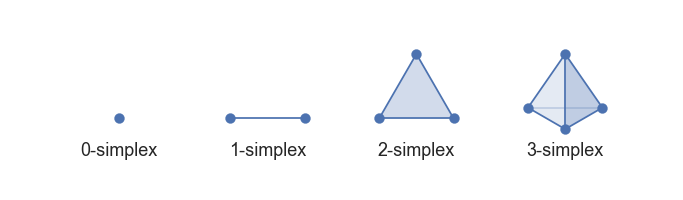} 
       \vspace{-0.5cm}
    \caption[Geometric representation of simplices]{\footnotesize Geometric representation of simplices\footnotemark. A collection of simplices makes up a simplicial complex which is used to approximate the shape structure underlying a discrete set of data points.}
   \label{fig:simplex}
   \vspace{-0.5cm}
\end{center}
\end{wrapfigure}

In TDA, the theory of persistent homology (PH) is one of the widely used tools to extract information about global shape, and local geometry of data \cite{zomorodian2005computing,edelsbrunner2008persistent}. Under this technique, we first construct on top of data points a nested sequence of (abstract) \emph{simplicial complexes} (called a \emph{filtration}) indexed by some suitable scale parameter. A simplicial complex consists of simplices (of various dimensions) as its building blocks. A 0-simplex can be identified with a node, 1-simplex with an edge, 2-simplex with a triangle, 3-simplex with a tetrahedron, and so on (see Figure \ref{fig:simplex}). A simplicial complex's dimension is defined as the maximum dimension of its simplices. From a practical point of view, simplicial complexes serve as a bridge between discrete data points and the underlying continuous shape they are sampled from \cite{nanda2014simplicial}. 

\footnotetext{Source: \url{https://umap.scikit-tda.org/how_umap_works.html}}

Various types of simplicial complexes exist, and selecting which one is appropriate to a given data set depends on the data structure, complexity, and size \cite{chazal2021introduction}. The \textit{Vietoris-Rips} complex is among the widely used simplicial complexes in applications. It can be efficiently constructed whenever there is a notion of distance between the data points \cite{ghrist2008barcodes}. Since there is no intrinsic notion of distance between the nodes of a graph, to construct the Vietoris-Rips filtration over it, the graph first must be embedded into a metric space. A straightforward way to achieve this is to have the distance between the nodes be the length of the shortest path (or geodesic distance) or be induced (if available) from edge weights representing node similarities (e.g., via pairwise correlations). However, under this approach, we lose the node connectivity information, and as the number of points increases, the size of the Vietoris-Rips complex grows exponentially. To curb a high computational cost for large graphs (e.g., of size 1000 nodes or more), in practice, one can terminate a Vietoris-Rips filtration much before the scale parameter reaches the maximum distance between the data points. On the downside, such a filtration can lead to a potential loss of vital topological information. Moreover, due to the so-called Nerve theorem, it is well known that a Vietoris-Rips complex generally does not recover the underlying shape structure at every value of the scale parameter \cite{nanda2014simplicial}. 

In the context of extracting topological information from a graph with (quantitative) node attributes, we instead adopt the \emph{lower-star} filtration \cite{edelsbrunner2022computational} due to its much lower computational cost and ability to preserve the edge connectivity structure. Both the lower-star and Vietoris-Rips approaches are examples of the \emph{sublevelset} filtration, which provides a general framework for constructing filtrations using sublevel sets of a monotone function $g$ defined over a simplicial complex \cite{nanda2014simplicial}. For example, the Vietoris-Rips filtration is the sublevel set filtration induced by $g(\tau)=\max_{u,v\in\tau}\{\rho(u,v)\}$, where $\rho(u,v)$ is the distance between the points $u$ and $v$ belonging to a simplex $\tau$. 

Now, we explain how lower-star filtration is defined. Given a graph $\mathcal G=(V,E;g)$, where $V$ and $E$ are the sets of nodes and edges of $\mathcal G$ respectively, and the values of $g:V\rightarrow\mathbb{R}$ represent node attributes. The dimension of $\mathcal G$ (viewed as a one-dimensional simplicial complex) can be increased by adding triangles (2-simplices), tetrahedrons (3-simplices), etc., formed by the graph nodes and edges. We continue to denote the enlarged simplicial complex for notational convenience by $\mathcal G$. A simple way to extend the domain of $g$ over the entire $\mathcal G$ is to use the maximum function: $g(\tau):=\max_{u\in \tau}\{g_v(u)\}$. It is easy to see that $g$ is monotone and hence induces a filtration of $\mathcal G$ called the lower-star filtration. 

Next, using the theory of persistence homology, we compute a \emph{persistence diagram} (PD) -- a multi-scale topological summary or a fingerprint of the data, which will be our main object to detect anomalous structural behaviors within a sequence of graphs ordered in time. The persistent homology allows us to study the quantitative topological changes along with filtration. More specifically, we track the dynamics of various topological features (such as connected components, loops, voids, etc.) by recording the instances when they first appear and later disappear as the scale parameter (by which the subcomplexes are indexed) increases. A table that stores this information is called a persistence diagram. Every row in a persistence diagram corresponds to a particular topological feature and contains its homological dimension (zero if a connected component, one if a loop, two if a void, etc.) and the scale values at which it appears and disappears. When a homological dimension is understood, we will write a persistence diagram with $n$ points as $D=\{(b_i,d_i)\}_{i=1}^n$, where $b_i$ and $d_i$ are the birth and death values of the $i$-th topological feature respectively. Since there could be more than one topological feature with the same birth and death values, a persistence diagram is a multi-set rather than simply a set of points in the plane.

\begin{wrapfigure}{r}{0.2\textwidth}
\vspace{-0.5cm}
  \begin{center}
    \includegraphics[width=0.15\textwidth]{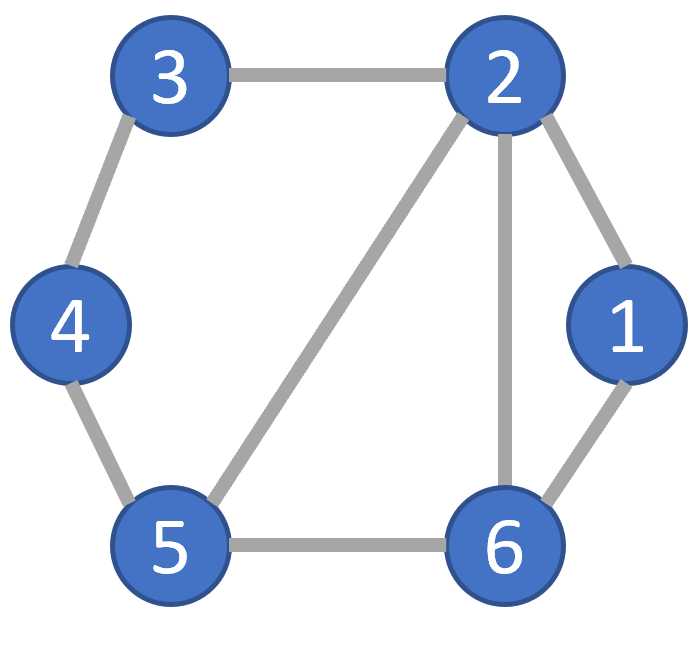}
    \caption{\footnotesize A simple toy graph $\mathcal G$ with six nodes and eight edges. We can view $\mathcal G$ as a one-dimensional simplicial complex with six 0-simplices (nodes) and eight 1-simplices (edges).}
    \label{fig:simple_graph}
  \end{center}
\vspace{-0.5cm}
\end{wrapfigure}

\begin{example}\label{ex:toy_graph}
In this example, we construct the Vietoris-Rips and lower-star filtrations for a simple graph. Let $\mathcal G=(V,E)$ be a graph with the set of nodes $V=\{1,2,3,4,5,6\}$ and edges $E=\{(1,2),(1,6),(2,3),(2,5),(2,6),(3,4),(4,5),(5,6)\}$ (see Figure~\ref{fig:simple_graph}). 
\end{example}

\begin{figure*}
\centering
\begin{subfigure}{.48\textwidth}
  \centering 
  \includegraphics[width=.95\linewidth]{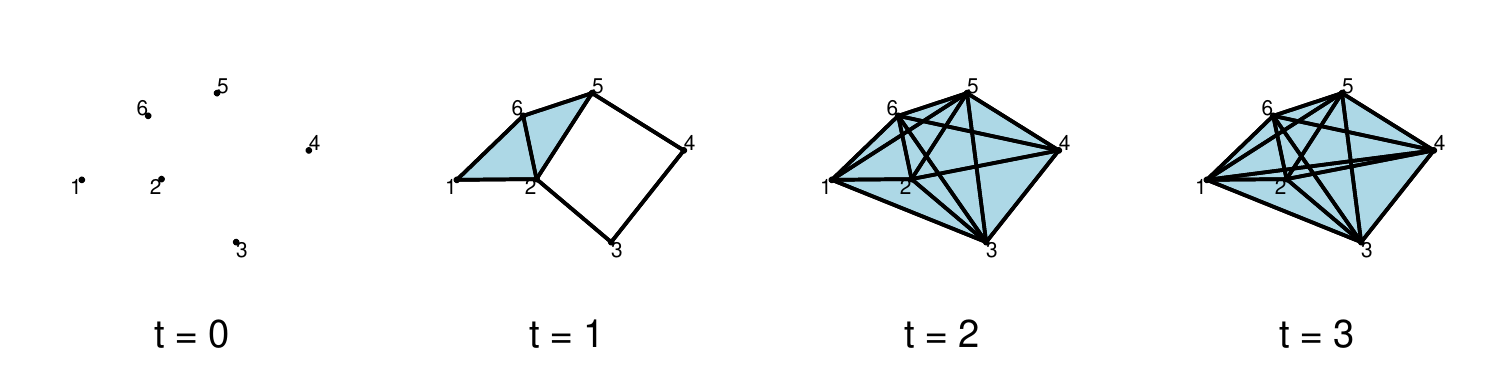}
  \caption{Vietoris-Rips filtration.}
  \label{fig:rips_filtration}
\end{subfigure}~
\begin{subfigure}{.48\textwidth}
  \centering 
  \includegraphics[width=.95\linewidth]{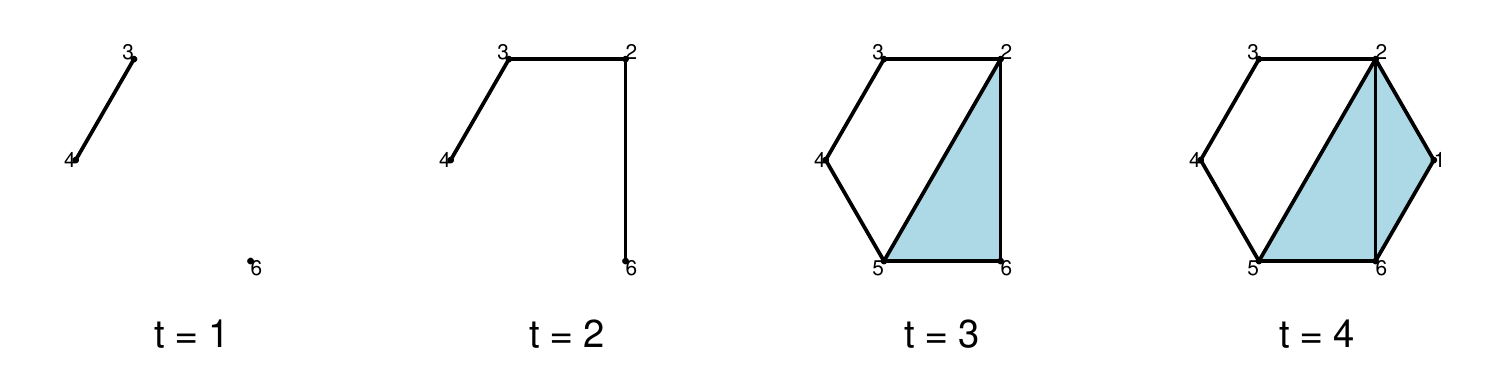}
  \caption{Lower-star filtration.}
  \label{fig:lowerstar_filtration}

\end{subfigure}
 \vspace{-0.2cm}
\caption{\footnotesize The Vietoris-Rips (VR) and lower-star (LS) filtrations of the toy graph $\mathcal{G}$ in Figure~\ref{fig:simple_graph}. To construct the VR filtration, the graph nodes are first embedded into a metric space using the geodesic distance. For the LS filtration, the dimension of $\mathcal G$, viewed as a one-dimensional simplicial complex, is first increased from one to two by adding the triangles formed by the graph edges. The resulting filtration is induced by function $g$ defined in Example \ref{ex:toy_graph} on the set of nodes of $\mathcal{G}$. Notice that the VR filtration involves a lot more 1-simplices (edges) and 2-simplices (shaded triangles) than the LS filtration.}
\vspace{-0.2cm}
\end{figure*}

We use the shortest path length between the nodes as a distance measure to embed the nodes of $\mathcal G$ into a metric space. Then we construct a two-dimensional Vietoris-Rips complex on top of the embedded points. The filtration of this complex is illustrated in Figure \ref{fig:rips_filtration}. The scale parameter $t$ represents the distance between the points in the embedded space. Initially, when $t=0$, there are six connected components and no loops. At $t=1$, several edges (1-simplices) and one triangle (2-simplex) are added, leading to the death of all but one connected component and the birth of a loop which later dies at $t=2$ (see the corresponding persistence diagram in Table \ref{PD:filtTables}). Observe that the number of 0-dimensional features in the persistence diagram is equal to the number of nodes in $\mathcal G$. 

For a lower-star filtration, we first construct a two-dimensional simplicial complex on top of $\mathcal G$ by adding 2-simplices (triangles). Next, to induce a filtration, we need to have a function $g$ defined on the set of nodes of $\mathcal G$. For example, suppose $g(1)=4$, $g(2)=2$, $g(3)=1$, $g(4)=1$, $g(5)=3$ and $g(6)=1$.  Figure \ref{fig:lowerstar_filtration} shows the lower-star filtration of $\mathcal G$ induced by $g$. At first, when $t=1$, two connected components are born, and later at $t=2$, one dies due to two edges (1-simplices) being added. The remaining connected component persists indefinitely. At $t=3$, three more edges and a triangle are added, resulting in the birth of the first loop. Since this loop never gets filled in by a triangle at a later point, it has an infinite death value. The PD of the lower-star filtration is given in Table \ref{PD:filtTables}. In practice, topological features with infinite death values can either be dropped or the death values be replaced by some suitable constant. 

\begin{wraptable}{l}{6.5cm}
\vspace{-0.25cm}
\vspace{-0.1cm}
 \centering
 \begin{tabular}{c c r c c}
\toprule
 \multicolumn{1}{c}{Filtration} & \multicolumn{1}{c}{Dim} & \multicolumn{1}{c}{Birth} & \multicolumn{1}{c}{Death} \\
 \midrule
 &0&0& $\infty$\\
 & 0 &0&1\\
 & 0 &0&1\\
VR  & 0 &0&1\\
 & 0 &0&1\\
 & 0 &0&1\\
 & 1 &1&2\\ 
 \midrule
 &0&1& $\infty$\\
LS & 0 &1&2\\
 & 1 &3&$\infty$\\
\bottomrule

 \end{tabular}
 	\caption{\footnotesize Persistence diagrams of the  Vietoris-Rips (VR) and lower-star (LS) filtrations of the toy graph $\mathcal G$ in Figure~\ref{fig:simple_graph}. Topological features of dimension 0 and 1 are connected components and loops respectively. See the depictions of the VR and LS filtrations in Figures \ref{fig:rips_filtration} and \ref{fig:lowerstar_filtration}.}
  
	\label{PD:filtTables}
\vspace{-1cm}
\end{wraptable}

A standard metric to measure the closeness between PDs is the Wasserstein distance. For technical reasons, PDs are assumed to additionally include the points of the diagonal $\Delta=\{(b,d)\in \mathbb{R}^2| b=d\}$ with infinite multiplicity.  
\begin{definition}
Let $D_1$ and $D_2$ be two persistence diagrams (of homological dimension $k$). The $L_p$ $q$-Wasserstein distance \cite{kerber2017geometry}, $p,q\geq 1$, is given by
$$
d_{pq}(D_1,D_2)=\Big(\inf_{\phi:D_1:\rightarrow D_2} \sum_{u\in D_1} \|u-\phi(u)\|_{p}^q \Big)^{1/q},
$$
where $\phi$ is a bijection such that an off-diagonal point is either matched with another off-diagonal point from the other diagram or with its own orthogonal projection on the diagonal. 
\end{definition}

When the two PDs have finitely many points, an optimal bijection that achieves the infimum always exists and can be found using the so-called Hungarian algorithm \cite{kuhn1955hungarian}. If $p=\infty$, the resulting distance is called the $q$-Wasserstein distance and if additionally $q=\infty$, the \emph{bottleneck} distance. 

\newpage

\subsection{A New Vectorization Scheme for Betti Functions and Stability Results} \label{subsec:VAB}
Since PDs do not form a Hilbert space \cite{bubenik2018topological,mileyko2011probability}, they can not be easily used in a wide range of predictive algorithms as input features of data. A common way to deal with this limitation is to extract functional summaries from PDs and discretize them to produce vectors in $\mathbb{R}^d$. Methods such as persistence landscapes and images belong to this category of summaries. In the present work, we work with a \emph{Betti function} \cite{chazal2021introduction} -- a univariate functional summary extracted from a PD, for its flexibility, lower computational cost and robustness (see Theorem \ref{thm:stabilityBetti}). 

\begin{definition}
Let $D$ be a persistence diagram (of homological dimension $k$), the Betti function (associated with $D$ and $k$) is defined as 
\begin{equation}\label{eqn:betti_function}
\beta(t)=\sum_{(b,d)\in D}w(b,d)\chi_{[b,d)}(t),    
\end{equation}
where $w:D\rightarrow \mathbb{R}$ is a weight function and $\chi_{[b,d)}(t)=1$ if $t\in[b,d)$ and 0 otherwise. 
\end{definition}

A common choice for the weight function $w$ is a non-decreasing function of the persistence value $d-b$. When $w$ is the identity function, $\beta(t)$ is called the $k$-th Betti number at $t$ and represents the count of ($k$-dimensional) holes (connected components if $k=0$, loops if $k=1$, voids if $k=2$ etc.) which are born before or at scale value $t$ but have not died yet. It is important to note that including additional point of the diagonal to a PD does not change the value of $\beta(t)$ since $\chi_{[a,b)}(t)=0$ whenever $a=b$.

For certain types of filtrations, PDs are known to be robust to input noise \cite{cohen2007stability,chazal2014persistence,cohen2010lipschitz}. In other words, a small change to data leads to a close PD (with respect to some chosen distance) to the one before the change. Naturally, stability is a desired property of any summary extracted from a PD. For example, associated stability results exist for functional summaries such as persistence landscapes and images. In Theorem \ref{thm:stabilityBetti} we establish a stability result for the Betti functions. First, we prove a technical lemma needed for our main theorem. 

\begin{lemma}\label{lemma:stability}
Let $w:\mathbb{R}^2\rightarrow \mathbb{R}$ be a bounded differentiable function whose partial derivatives are also bounded. Then for any $u=(a,b)$ and $v=(c,d)$ with $a\leq b$ and $c\leq d$, we have
\begin{equation*}
    \int_{\mathbb{R}}|w(u)\chi_{[a,b)}(t)-w(v)\chi_{[c,d)}(t)|dt \\ \leq  (\|w\|_\infty+L\|\nabla w\|_\infty)\|u-v\|_1,
\end{equation*}
where $L\geq \max\{b-a,d-c\}$,
\begin{equation*}
   \|w\|_\infty=\sup_{z\in \mathbb{R}^2}|w(z)|\quad \hbox{and}\quad \|\nabla w\|_\infty=\sup_{z\in \mathbb{R}^2}\|\nabla w(z)\|_2 
\end{equation*}
\end{lemma}
\vspace{-0.5cm}
\begin{proof}
We will prove the lemma considering three separate cases. Let $g(t)=|w(u)\chi_{[a,b)}(t)-w(v)\chi_{[c,d)}(t)|$.
Case 1: $[a,b)\cap [c,d)\neq \emptyset$, $[a,b)\not\subseteq [c,d)$ and $[c,d)\not\subseteq [a,b)$. Without the loss of generality, assume $a\leq c\leq b \leq d $. Then
    \begin{align*}
        \int_{\mathbb{R}}g(t)dt &= \int_a^c g(t)dt+\int_c^b g(t)dt+\int_b^d g(t)dt\\
        &\leq \|w\|_\infty(|c-a|+|d-b|)+\|\nabla w\|_\infty\|u-v\|_2|b-c|\\
        &\leq \|w\|_\infty (|c-a|+|d-b|)+\|\nabla w\|_\infty \|u-v\|_1 L \\
        &=(\|w\|_\infty+L\|\nabla w\|_\infty)\|u-v\|_1.
\end{align*}

Case 2: $[a,b)\cap [c,d)= \emptyset$. Without the loss of generality, assume $a\leq b\leq c \leq d $. Then
    \begin{align*}
        \int_{\mathbb{R}}g(t)dt & = \int_a^b g(t)dt+\int_c^d g(t)dt\\
        &\leq \|w\|_\infty|b-a|+\|w\|_\infty|d-c|\\
        &\leq \|w\|_\infty (|c-a|+|d-b|)\\
        &=\|w\|_\infty\|u-v\|_1.
\end{align*}

Case 3: $[a,b)\subseteq [c,d)$ or $[c,d)\subseteq [a,b)$. Without the loss of generality, assume $a\leq c\leq d \leq b $. Then
    \begin{align*}
        \int_{\mathbb{R}}g(t)dt &= \int_a^c g(t)dt+\int_c^d g(t)dt+\int_d^b g(t)dt\\
        &\leq \|w\|_\infty(|c-a|+|b-d|)+\|\nabla w\|_\infty\|u-v\|_2|d-c|\\
        &\leq \|w\|_\infty (|c-a|+|b-d|)+\|\nabla w\|_\infty \|u-v\|_1 L\\
        &=(\|w\|_\infty+L\|\nabla w\|_\infty)\|u-v\|_1.
\end{align*}
The lemma follows by combining these three cases. 
\end{proof}

\begin{theorem}\label{thm:stabilityBetti}
Let $D_1$ and $D_2$ be two (finite) persistence diagrams with finite death values and let $\phi$ be an optimal bijection between them with respect to the $L_1$ 1-Wasserstein distance $d_{11}$. Denote by
$\beta_i(t)=\sum_{(b,d)\in D_i}w(b,d)\chi_{[b,d)}(t)$ the Betti function (associated with $D_i$, $i=1,2$) with weight function $w$ is as in Lemma \ref{lemma:stability}. Then
\[
\|\beta_1-\beta_2\|_{L_1}\leq (\|w\|_\infty+L\|\nabla w\|_\infty)d_{11}(D_1,D_2),
\]
where $L=\max_{(b,d)\in D_1\cup D_2}(d-b)$. In particular, if $w\equiv1$ then $\|w\|_\infty=1$, $\|\nabla w\|_\infty=0$ and therefore, 
\[
\|\beta_1-\beta_2\|_{L_1}\leq d_{11}(D_1,D_2).
\]
\end{theorem}
\vspace{-0.4cm}
\begin{proof}
 
Let $u=(b_u,d_u)$ and $\phi(u)=(b_{\phi(u)},d_{\phi(u)})$. Then
\begin{align*}
    \|\beta_1-\beta_2\|_{L_1}&=\int_{\mathbb{R}} \Big|\sum_{u\in D_1}\Big[w(u)\chi_{[b_u,d_u)}(t)-w(\phi(u))\chi_{ [b_{\phi(u)},d_{\phi(u)})}(t)\Big] \Big|dt\\
    &\leq \sum_{u\in D_1} \int_{\mathbb{R}} \Big|w(u)\chi_{[b_u,d_u)}(t)-w(\phi(u))\chi_{[b_{\phi(u)},d_{\phi(u)})}(t)\Big|dt\\
    &\leq  (\|w\|_\infty+L\|\nabla w\|_\infty) \sum_{u\in D_1} \|u-\phi(u)\|_1 \mbox{  \quad (by Lemma \ref{lemma:stability})}\\
    &= (\|w\|_\infty+L\|\nabla w\|_\infty)d_{11}(D_1,D_2),
\end{align*}

\end{proof}

\begin{figure}
\centering
\begin{subfigure}{.4\textwidth}
  \centering 
  \includegraphics[scale=0.27]{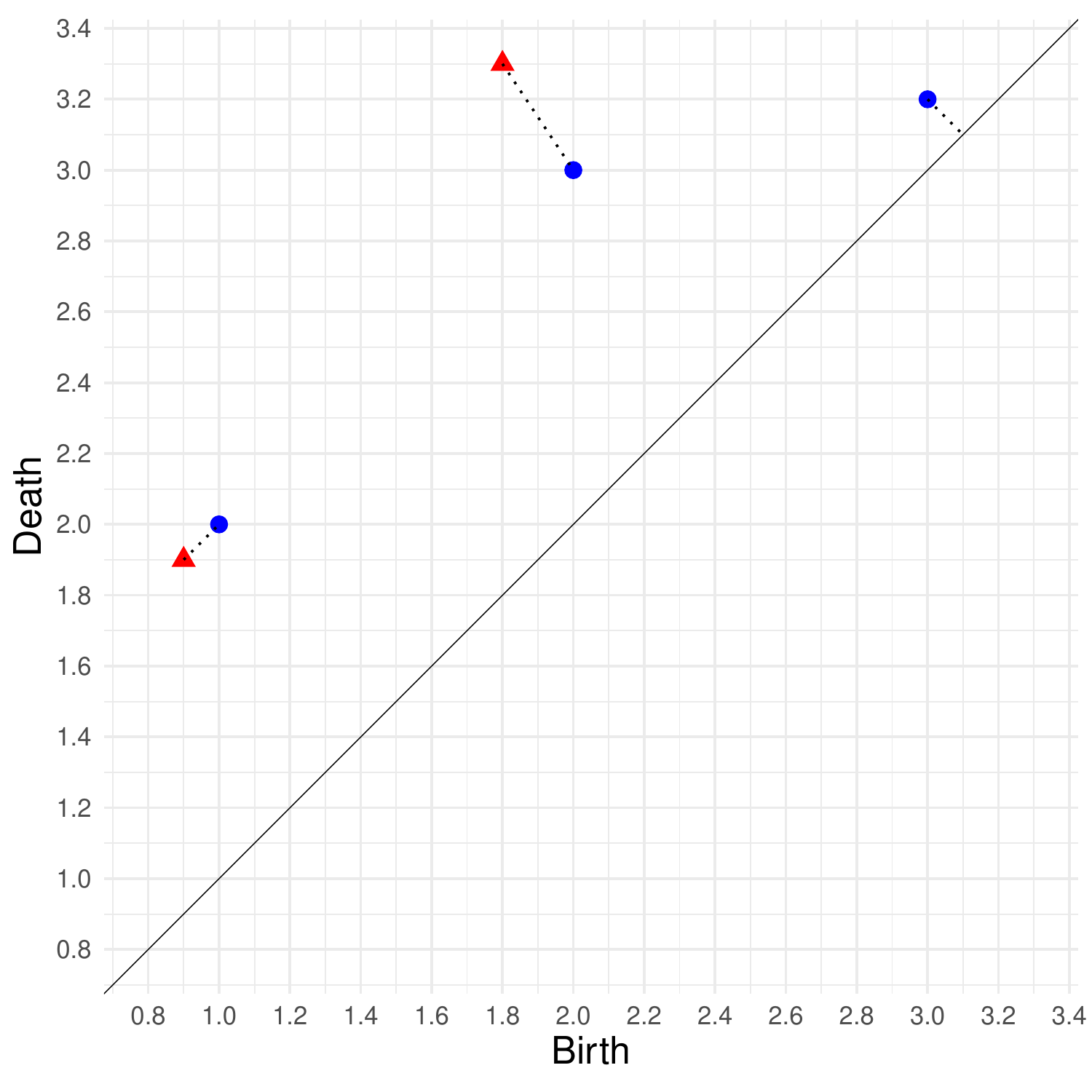}
	\vspace{-0.25cm}
	\caption{\footnotesize Plots of persistence diagrams $D_1$ (blue) and $D_2$ (red). The points matched by a optimal bijection of the $L_1$ 1-Wasserstein distance are joined by dashed lines. The blue point in the top-right corner is matched to a point on the diagonal. } 
	\label{fig:wasserstein}
\end{subfigure}\quad\quad\quad
\begin{subfigure}{.4\textwidth}
  \centering 
  \includegraphics[scale=0.42]{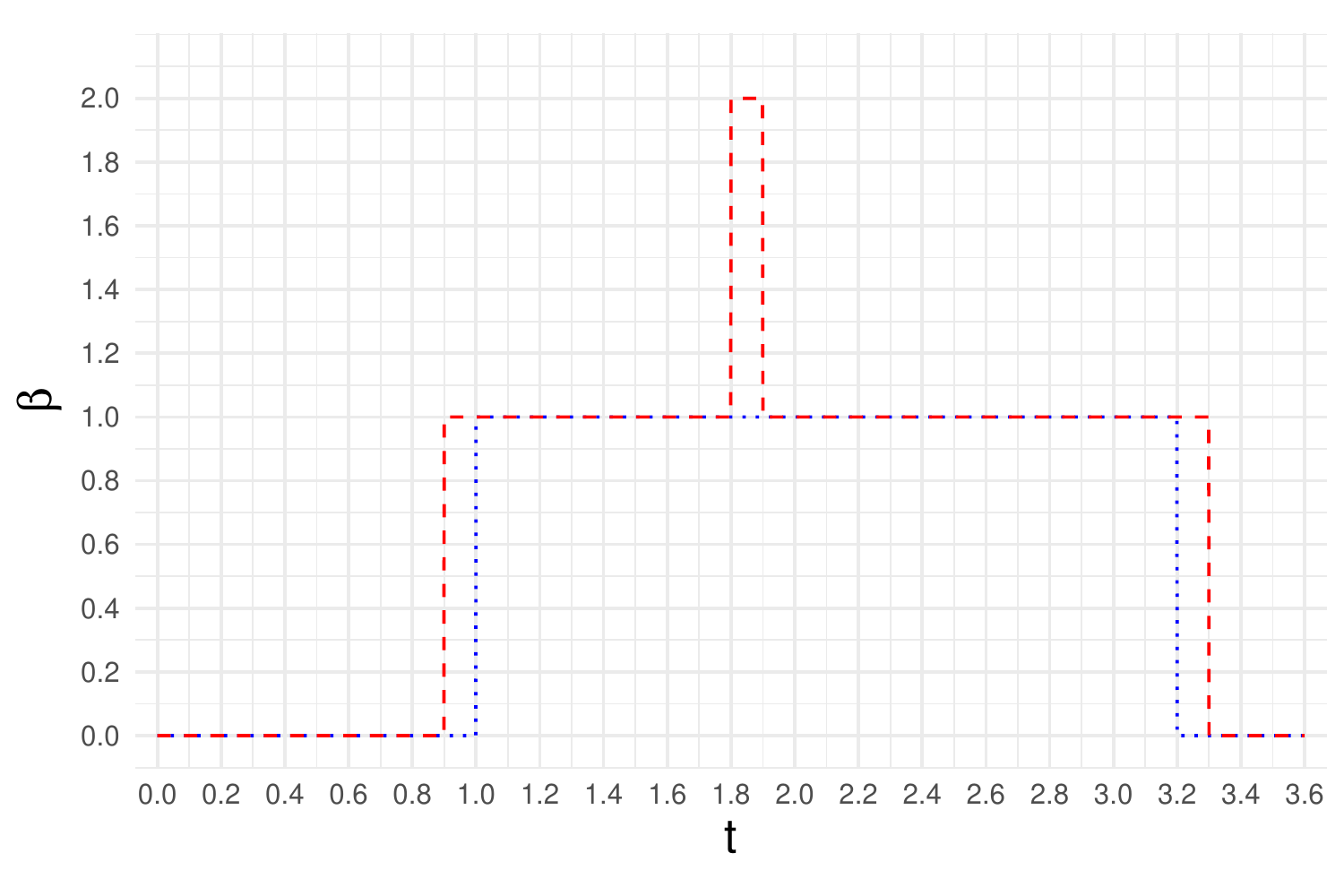}
	\vspace{-0.25cm}
	\caption{\footnotesize Graphs of Betti functions $\beta_1(t)$ (blue) and $\beta_2(t)$ (red) associated with the persistence diagrams $D_1$ and $D_2$ respectively. The weight function $w$ used in the construction of the Betti functions is identically equal to one.}
	\label{fig:betti}
\end{subfigure}
 \vspace{-0.1cm}
\caption{\footnotesize Plots and graphs for Example \ref{ex:matching}.}
\end{figure}

\begin{example}\label{ex:matching}
Let $D_1=\{(1,2),(2,3),(3,3.2)\}$ and $D_2=\{(0.9,1.9),(1.8,3.3)\}$ be two persistence diagrams. In Figure \ref{fig:wasserstein}, the points of $D_1$ and $D_2$ are depicted as blue circles and red triangles respectively. With respect to the $L_1$ 1-Wasserstein distance, the optimal bijection $\phi$ matches $(1,2)$ with $(0.9,1.9)$, $(2,3)$ with $(1.8,3.3)$ and $(3,3.2)$ with $(3.1,3.1)$ (shown with dashed lines in Figure \ref{fig:wasserstein}). Therefore,
\begin{align*}
    d_{11}(D_1,D_2)&=\|(1,2)-(0.9,1.9)\|_{1}+\|(2,3)-(1.8,3.3)\|_{1}+\|(3,3.2)-(3.1,3.1)\|_{1}\\
    &=\|(0.1,0.1)\|_{1}+\|(0.2,\text{-} 0.3)\|_{1}+\|(\text{-}0.1,0.1)\|_{1}\\
    &=|0.1|+|0.1|+|0.2|+|\text{-}0.3|+|\text{-}0.1|+|0.1|=0.9
\end{align*}
The associated Betti functions with $w(b,d)\equiv 1$ are $\beta_1(t)=\chi_{[1,2)}(t)+\chi_{[2,3)}(t)+\chi_{[3,3.2)}(t)=\chi_{[1,3.2)}(t)$ and $\beta_2(t)=\chi_{[0.9,1.9)}(t)+\chi_{[1.8,3.3)}(t)$ (see Figure \ref{fig:betti}). The $L_1$-distance between the two Betti functions can be calculated as
\begin{align*}
    \|\beta_1-\beta_2\|_{L_1}&=\int_{0.9}^1|\beta_1(t)-\beta_2(t)|dt+\int_{1.8}^{1.9}|\beta_1(t)-\beta_2(t)|dt+\int_{3.2}^{3.3}|\beta_1(t)-\beta_2(t)|dt\\
    &=\int_{0.9}^1|\chi_{[1,3.2)}(t)-\chi_{[0.9,1.9)}(t)|dt+\int_{1.8}^{1.9}|\chi_{[1,3.2)}(t)-\chi_{[0.9,1.9)}(t)-\chi_{[1.8,3.3)}(t)|dt\\
    & \quad + \int_{3.2}^{3.3}|-\chi_{[1.8,3.3)}(t)|dt=0.1+0.1+0.1=0.3
\end{align*}
\end{example}

A common way to discretize a Betti function $\beta(t)$ is to evaluate it at increasing points $t_1,t_2,\ldots,t_d$ and to arrange these values into a vector:
$$
\boldsymbol{\beta}^{\hbox{\tiny{com}}}=(\beta(t_1),\beta(t_2),\ldots,\beta(t_d)).
$$
In applications, one may have to choose $d$ large enough, e.g., 100, in order to preserve the functional shape of $\beta(t)$ in the resulting vector. We propose an alternative vectorization scheme which averages the Betti function between two adjacent points $t_i$ and $t_{i+1}$ via integration for each $i=1,2,\ldots,d-1$:
\begin{equation}\label{VAB}
\boldsymbol{\beta}=\Big(\frac{1}{\Delta t_1}\int_{t_1}^{t_2}\beta(t)dt,\ldots,\frac{1}{\Delta t_{d-1}}\int_{t_{d-1}}^{t_d}\beta(t)dt\Big)
\end{equation}
where $\Delta t_k=t_{k+1}-t_k$, $k=1,\ldots,d-1$. We call $\boldsymbol{\beta}$ a \emph{vector of averaged Bettis} (VAB). Due to simplicity of the Betti function in term of construction, the integrals in (\ref{VAB}) can be analytically computed and efficiently implemented in computer software. In the following theorem, we prove a stability result for VABs as a corollary of Theorem \ref{thm:stabilityBetti}.

\begin{corollary}\label{cor:stabilityVAB}
Let $D_1$ and $D_2$ be two (finite) persistence diagrams, and let $\phi$ be an optimal bijection between them with respect to the $L_1$ 1-Wasserstein distance. For a given increasing sequence of equally spaced scale values $t_1,t_2,\ldots,t_d$, let $\boldsymbol{\beta}_1$ and $\boldsymbol{\beta}_2$ denote the corresponding VABs defined in (\ref{VAB}) with the weight function $w$ as in Lemma \ref{lemma:stability}. Then
$$
    \|\boldsymbol{\beta}_1-\boldsymbol{\beta}_2\|_{1}\leq \frac{1}{\Delta t}(\|w\|_\infty+L\|\nabla w\|_\infty)d_{11}(D_1,D_2),
$$
where $\Delta t=t_{i+1}-t_i=\hbox{const}$. In particular, if $w\equiv1$ then, 
$$
    \|\boldsymbol{\beta}_1-\boldsymbol{\beta}_2\|_{1}\leq \frac{1}{\Delta{t}}d_{11}(D_1,D_2).
$$
\end{corollary}
\vspace{-0.5cm}
\begin{proof}
    \begin{align*}
        \|\boldsymbol{\beta}_1-\boldsymbol{\beta}_2\|_{1}&=\sum_{i=1}^{d-1}|\beta_{i1}-\beta_{i2}|=\sum_{i=1}^{d-1}\frac{1}{\Delta t}\Big|\int_{t_i}^{t_{i+1}}[\beta_1(t)-\beta_2(t)]dt \Big|\\
        &\leq \sum_{i=1}^{d-1}\frac{1}{\Delta t}\int_{t_i}^{t_{i+1}}\Big|\beta_1(t)-\beta_2(t)\Big|dt\leq \frac{1}{\Delta t}\int_{\mathbb{R}}\Big|\beta_1(t)-\beta_2(t)\Big|dt\\
        &=\frac{1}{\Delta t}\|\beta_1-\beta_2\|_{L_1}\leq \frac{1}{\Delta t}(\|w\|_\infty+L\|\nabla w\|_\infty)d_{11}(D_1,D_2) \quad \hbox{ (by Theorem \ref{thm:stabilityBetti})}
    \end{align*}
\end{proof}


\section{Experimental Results}
\label{sec:experiments}
\subsection{Change-point Detection in Time-varying Simulated Graphs}
In this simulation study, we compare the two approaches to vectorize a Betti function in the context of a change-point detection problem when the extracted topological vector summaries are low-dimensional (in higher dimensions, both vectorizations tend to yield the same results). We further compare the performance of the topological features\footnote{In Section \ref{sec:experiments} and \ref{sec:conclusion}, by topological features we mean topological input variables (or predictors) and not the features such as connected components, loop, voids, etc. mentioned in Section \ref{sec:theory}.} to two sets of graph features. The first is a 2-vector whose elements are the counts of 3-motifs (connected subgraphs of size three) with two and three edges ($\angle$ and $\triangle$). The second set of features includes standard graph summary statistics such as the number of edges, the global clustering coefficient, the associativity coefficient, and the degree of centralization. The global clustering coefficient, also known as the global graph transitivity, is three times the ratio of the number of triangles and the connected subgraphs of size three. The assortativity coefficient of a graph quantifies the tendency of similar nodes to interconnect. As for the degree of centralization, it measures the extent to which the nodes compare to the most central node in terms of the number of connections. 

\subsubsection{Simulation Settings}

The procedure we use to generate graphs and compute topological summaries is given as follows:
\begin{enumerate}
    \item Sample $n$ points from the Dirichlet distribution with a given parameter $\alpha\in\mathbb{R}^m$. Each sample point $x=(x_1,\ldots,x_m)$ is a vector whose elements are non-negative numbers adding up to one and hence can be viewed as probabilities.  
    \item Generate a graph of size $n$ where the nodes correspond to the sampled points and the edge between nodes $x$ and $y$ appears with probability equal to the dot product between $x$ and $y$, i.e., $x\cdot y\in [0,1]$. Such a graph is called a \emph{random dot product graph} which has been used to model social networks \cite{kraetzl2005random}.
    \item Enlarge the graph to a two-dimensional simplicial complex by adding the triangles (2-simplices) formed by the graph edges. 
    \item Use the cross-entropy function to produce node attributes: $g(x)=-\sum_{i=1}^m x_i\log(x_i)$ for  node $x$.
    \item Compute the PD of the lower-star filtration induced by $g$ (whose values are normalized to lie in $[0,1]$).
    \item Compute  $\boldsymbol{\beta^{\hbox{\tiny{com}}}}$ and $\boldsymbol{\beta}$ taking $d$ equally spaced points in $[0,1]$ to vectorize the associated Betti function for homological dimensions 0 and 1. 
\end{enumerate}

Following the above procedure, we choose $m=3$ and fix $\alpha_1=(1.5,1.5,1.5)$, $\alpha_2=(2,2,2)$, $\alpha_4=(2,2,3.5)$ and $\alpha_4=(2,0.5,2)$, and generate graphs $\mathcal G_{t,\alpha}$ where $\alpha=\alpha_{\lceil t/50\rceil}$ and $t=1,\ldots,200$. Thus, there are 50 graphs for each choice of $\alpha$ and the change points occur at points $t=51$, $t=101$ and $t=151$. 

Since topological vector summaries are generally not known to follow a parametric distribution, we resort to a non-parametric approach to estimate the change points. More specifically, we use a permutation-based hierarchical algorithm for multiple change point detection called E-divisive \cite{matteson2014nonparametric}. It estimates change points by applying a permutation-based iterative procedure to detect a single change point. For each homological dimension (i.e., 0 and 1), we input the corresponding sequences $\{\boldsymbol{\beta}_{t,\alpha}^{\hbox{\tiny{com}}}\}$ and $\{\boldsymbol{\beta}_{t,\alpha}\}$, computed for $d=5$, to the E-divisive algorithm which returns a list of estimated change points. Let $\{\boldsymbol{m}_{t,\alpha}\}=\{(M_{\angle,t,\alpha},M_{\triangle,t,\alpha})\}$ be a sequence of the counts of the two 3-motifs and $\{\boldsymbol{g}_{t,\alpha}\}=\{(|E_{t,\alpha}|,C_{t,\alpha},r_{t,\alpha},DC_{t,\alpha})\}$ denote a sequence of vectors of the above standard graph summaries which are also inputted to the change-point detection algorithm. 

\subsubsection{Performance Results}

We use the absolute error between a true change point and its estimate as the performance evaluation metric. If E-divisive does not detect the expected change point, the absolute error is set to 50, which is the distance between any two nearest change points. Table \ref{table:CPD} shows the mean absolute errors (MAE) (computed over 100 independent simulation trials) associated with the estimated change points for each feature input. 

The fact that $\alpha_{2k}/\alpha_{1k}=4/3$ for all $k=1,2,3$ results in the distributions $\hbox{Dir}(\alpha_1)$ and $\hbox{Dir}(\alpha_2)$ having the same marginal means. In this case, the differences in the marginal variances are all less than 0.009, which leads to insignificant changes in the motif and graph summaries (see Table \ref{table:summaries}) and hence a poor detection rate of the first change point. On the other hand, both topological summaries show enough sensitivity to detect the transition from $\alpha_1$ to $\alpha_2$. 
\begin{table}[t]
\vspace{0.2cm}
 	\centering
	\begin{tabular}{lcccc}
		\toprule
		Input & 1st CP & 2nd CP & 3rd CP \\ 
		\toprule
		Common Betti (dim=0) & 5.05 & 38.64 & 0.91  \\ 
		Common Betti (dim=1) & 1.08 & 1.26 & 0.87  \\
		VAB (dim=0) & 2.28 & 4.31 & 0.22  \\
		VAB (dim=1) & 2.47 & 0.32 & 0.17  \\
		Graph summaries & 48.69 & 0.07 & 0.07  \\
		Motifs & 47.82 & 0.07 & 0.11  \\
		\bottomrule
	\end{tabular}
		 \vspace{0.2cm}
	\caption{\footnotesize Mean absolute errors (computed over 100 simulation runs) of the estimated change points (CP) in time-varying graphs indexed by $t\in \{1,2,\ldots,200\}$. The graphs are generated according to the random dot product graph model. The true CPs occur at the time points $t=51$, $t=101$ and $t=151$ when the graph generating process undergoes a distributional change. The CPs are estimated using a permutation-based algorithm called E-divisive \cite{matteson2014nonparametric}.}
	\label{table:CPD}
	\vspace{-0.3cm}	
 \end{table} 
 
On the second change point, the relative change in the mean of $\|\boldsymbol{\beta}_0^{\hbox{\tiny{com}}}\|$ is 7\% and the corresponding MAE is quite high (38.64). Whereas for $\|\boldsymbol{\beta}_0\|$, this relative change is 14\%, and consequently, the MAE is comparatively much lower (4.31). When switching from $\alpha_2$ to $\alpha_3$ and $\alpha_3$ to $\alpha_4$, both marginal means and variances of the Dirichlet distributions undergo changes which lead to more pronounced changes in the employed summaries and hence to higher detection rates. According to Table \ref{table:CPD}, using VABs results in smaller MAEs compared to those produced by the competing feature summaries. 

\begin{table*}[t]
\centering
\begin{tabular}{ccccccccccc}
\toprule
 & $M_\angle$ & $M_\triangle$ & $|E|$ & $C$ & $r$ & $DC$ & $\|\boldsymbol{\beta}_0^{\hbox{\tiny{com}}}\|$ & $\|\boldsymbol{\beta}_1^{\hbox{\tiny{com}}}\|$ & $\|\boldsymbol{\beta}_0\|$ & $\|\boldsymbol{\beta}_1\|$ \\ 
\toprule
$\alpha_1$ & 35736.69 & 6069.28 & 1650.89 & 0.34 & -0.02 & 0.12 & 2.59 & 4.63 & 2.18 & 5.93 \\ 
  $\alpha_2$ & 35765.30 & 6014.05 & 1648.44 & 0.33 & -0.02 & 0.12 & 2.01 & 6.60 & 1.94 & 5.09 \\ 
  $\alpha_3$ & 39867.43 & 7696.07 & 1776.06 & 0.37 & 0.01 & 0.14 & 1.86 & 1.62 & 1.66 & 1.40 \\ 
  $\alpha_4$ & 47268.79 & 11527.49 & 2021.17 & 0.42 & 0.00 & 0.14 & 2.16 & 1.26 & 2.02 & 1.23 \\ 
\bottomrule
\end{tabular}
	\vspace{-0.1cm}
	 \caption{\footnotesize Means of the motif, graph and topological summaries over 100 simulation runs for different vectors of the $\alpha$ parameter for the Dirichlet distribution. A change point occurs whenever there is a transition from $\alpha_k$ to $\alpha_{k+1}$, $k=1,2,3$, in the distribution. $\|\cdot\|$ denotes the Euclidean norm.}
	\label{table:summaries}
	\vspace{-0.05cm}
\end{table*}
  
\subsection{Anomalous Price Prediction in the Ethereum Transaction Networks}



Vitalik Butterin and Gavin Wood created the Ethereum blockchain in 2015~\cite{wood2014ethereum}. Ethereum is a blockchain platform because Ethereum's primary goal is to store data and software code (called smart contracts) on a blockchain. The code storage and execution on platforms is the greatest difference between cryptocurrencies and platforms. Similar to Bitcoin, Ethereum has a cryptocurrency: Ether. Additionally, any Ethereum user can deploy a smart contract to create several digital assets, such as ERC-20 and ERC-721 tokens. The asset, called a smart contract-based token, has a total supply where each instance can be traded among ethereum addresses.   

In this study, we analyze daily transaction graphs (or networks) of Ethereum tokens and aim to predict anomalous price behaviors. Using the tool ethereum-etl,\footnote{\url{https://github.com/blockchain-etl/ethereum-etl}} we extract daily transfers of 31 tokens during a period from May 2017 to May 2018. For each token transaction, the data provides the sender and receiver addresses, the amount, and the date it is transferred on. We normalize the transaction amounts per token using the log transformation and consider only those days on which at least five tokens are traded. After these preprocessing steps, the final data set contains about 10 million transactions. 


Let $\mathcal{G}_t$ be the token transaction graph for day $t\in \{1,2,\ldots,T\}$ and $P_t$ the corresponding Ethereum token open price. Define the binary response anomalous price variable, $Y_{t}$, by 
\begin{equation*}
Y_{t} = \left\{
        \begin{array}{ll}
            1 \hbox{ (anomalous)} & \hbox{ if }  |R_t|\geq\delta \\
            0 \hbox{ (normal)} & \hbox{ otherwise},
        \end{array}
    \right.
\end{equation*}
where $R_t = (P_{t} - P_{t-1})/(P_{t-1})$ is the open price return on day $t$ and  $\delta>0$ a user-defined threshold for the magnitude of a price shock. For each day $t$, the goal is to predict if there will be a price shock in the next $h$ days (prediction horizon) i.e., if $\exists s\in \{t+1,\ldots,t+h\}$ such that $Y_{s}=1$ based on a set of feature variables describing the transaction graph $\mathcal{G}_t$ including the normalized price $PN_t=P_t/\max\{P_1, \ldots, P_{T}\}$, the number of edges $|E_t|$, number of nodes $|V_t|$ and the average clustering coefficient $C_t$.  

Now, we explain a two-step procedure to define topological features to consider along with the others. First, we take the sum of mean amounts sent and received per node as the node attribute. Quantitative node attributes are needed to be able to extract topological information from $\mathcal{G}_t$ using the lower-star filtration. To reduce the computational cost, we trim $\mathcal{G}_t$ keeping only the most active $M$ nodes. Next, we upgrade $\mathcal{G}_t$ to a two-dimensional simplicial complex by adding triangles (2-simplices) formed by its edges and compute the associated vector of averaged Bettis (VAB) $\boldsymbol\beta_{k,t}$ of length 99 ($d=100$), where $k\in\{0,1\}$ is the homological dimension. Here, we seek to capture the functional form of the underlying Betti functions by choosing $d=100$ (versus $d=5$ as in the previous experiment). 

Instead of supplying the $d-1=99$ features provided by the VABs to a model, in the second step of the procedure, we define new (scalar) topological features employing the concept of functional data depth. More specifically, we use the modified band depth (MBD) \cite{lopez2009concept} to measure the "outlyingness" of $\boldsymbol\beta_{k,t}$ with respect to those of the past $w$ days (including the present-day $t$). Define the \emph{rolling depth} of $\boldsymbol\beta_{k,t}$ as
\begin{equation*}
RD_w(\boldsymbol\beta_{k,t})= MBD(\boldsymbol\beta_{k,t}| \boldsymbol\beta_{k,t},\boldsymbol\beta_{k,t-1},\ldots\boldsymbol\beta_{k,t-w+1})=\binom{w}{2}^{-1}\sum_{0\leq s < m\leq w-1}\frac{1}{d-1}\sum_{i=1}^{d-1}\chi_{(a_i,b_i)}(\boldsymbol\beta_{k,t,i}),
\end{equation*}
where $\boldsymbol\beta_{k,t,i}$ is the $i$-th element of the vector $\boldsymbol\beta_{k,t}$, $a_{i}=\min\{\boldsymbol\beta_{k,t-s,i},\boldsymbol\beta_{k,t-m,i}\}$, $b_{i}=\max\{\boldsymbol\beta_{k,t-s,i},\boldsymbol\beta_{k,t-m,i}\}$. Clearly, $RD_w(\boldsymbol\beta_{k,t})\in[0,1]$ and a value close to 1 (maximum depth value) would indicate a similar behavior (in terms of the graph's topology) on day $t$ relative to the recent $w$ days. On the other hand, a value of $RD_w$ near the minimum depth value of 0 signals an anomalous situation (a price shock) on day $t$ with respect to the last $w$ days.  

Finally, we fit a random forest model to three sets of features. An optimal number of features randomly sampled at each split is selected using the 5-fold cross-validation. Model $M_1$ is the baseline model without any topological input. Model $M_2$ has one additional feature given by $RD_7(\boldsymbol\beta_{0})$ and $M_3$ is the full model containing all of the available features (see Table \ref{table:models}).  

\begin{table*}[t]
\centering
	\begin{tabular}{cll}
	   \toprule
		\bf{Model} & \bf{Description} &\bf{Features} \\ 
		\midrule
		$M_1$  & Baseline & $PN$, ${|E|}$, $|V|$, $C$ \\  
		$M_2$  & Baseline + 0-dim top. features &$PN$, ${|E|}$, $|V|$, $C$,  $RD_7(\boldsymbol\beta_{0})$ \\  
		$M_3$  & Baseline + 0 \& 1-dim top. features &$PN$, ${|E|}$, $|V|$, $C$,  $RD_7(\boldsymbol\beta_{0}), RD_7(\boldsymbol\beta_{1})$\\  
	\bottomrule
	\end{tabular}
	 \vspace{-0.15cm}
	\caption{\footnotesize Three sets of features used as input variables for modelling. $PN$ - normalized price, $|E|$ - number of edges, $|V|$ - number of nodes, $RD_7(\boldsymbol\beta_{0})$, $RD_7(\boldsymbol\beta_{1})$ - rolling depth values of $\boldsymbol\beta_{0}$ and $\boldsymbol\beta_{1}$ with respect to past 7 days.} 
	\label{table:models}
\end{table*}

\begin{figure*}[!t]
\centering
\begin{subfigure}{.32\textwidth}
  \centering 
  \includegraphics[width=.95\linewidth]{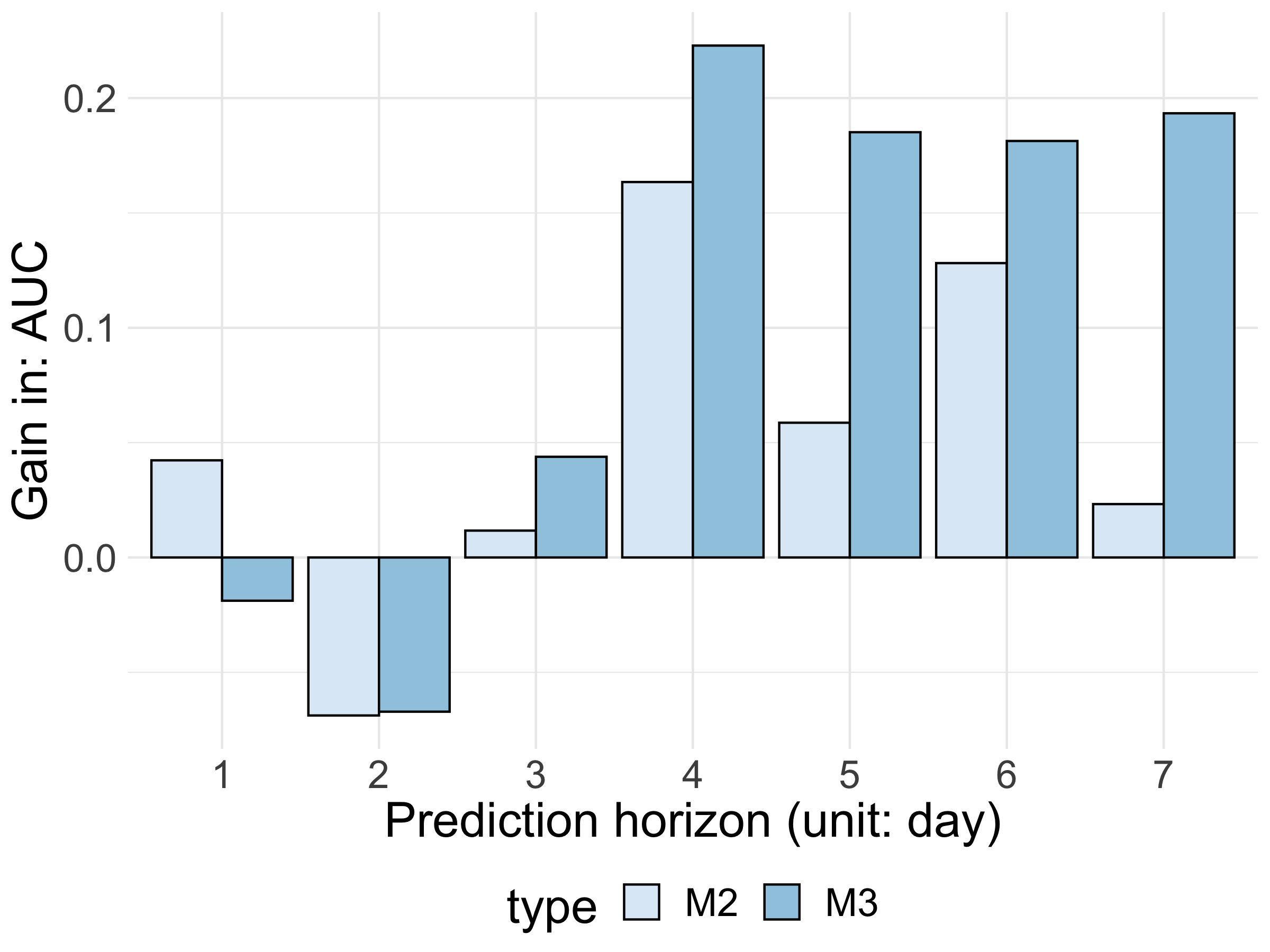}
  \caption{Gains in AUC for VABs with respect to baseline model $M_1$.}
  \label{fig:AUC_betti}
\end{subfigure}~
\begin{subfigure}{.32\textwidth}
  \centering 
  \includegraphics[width=.95\linewidth]{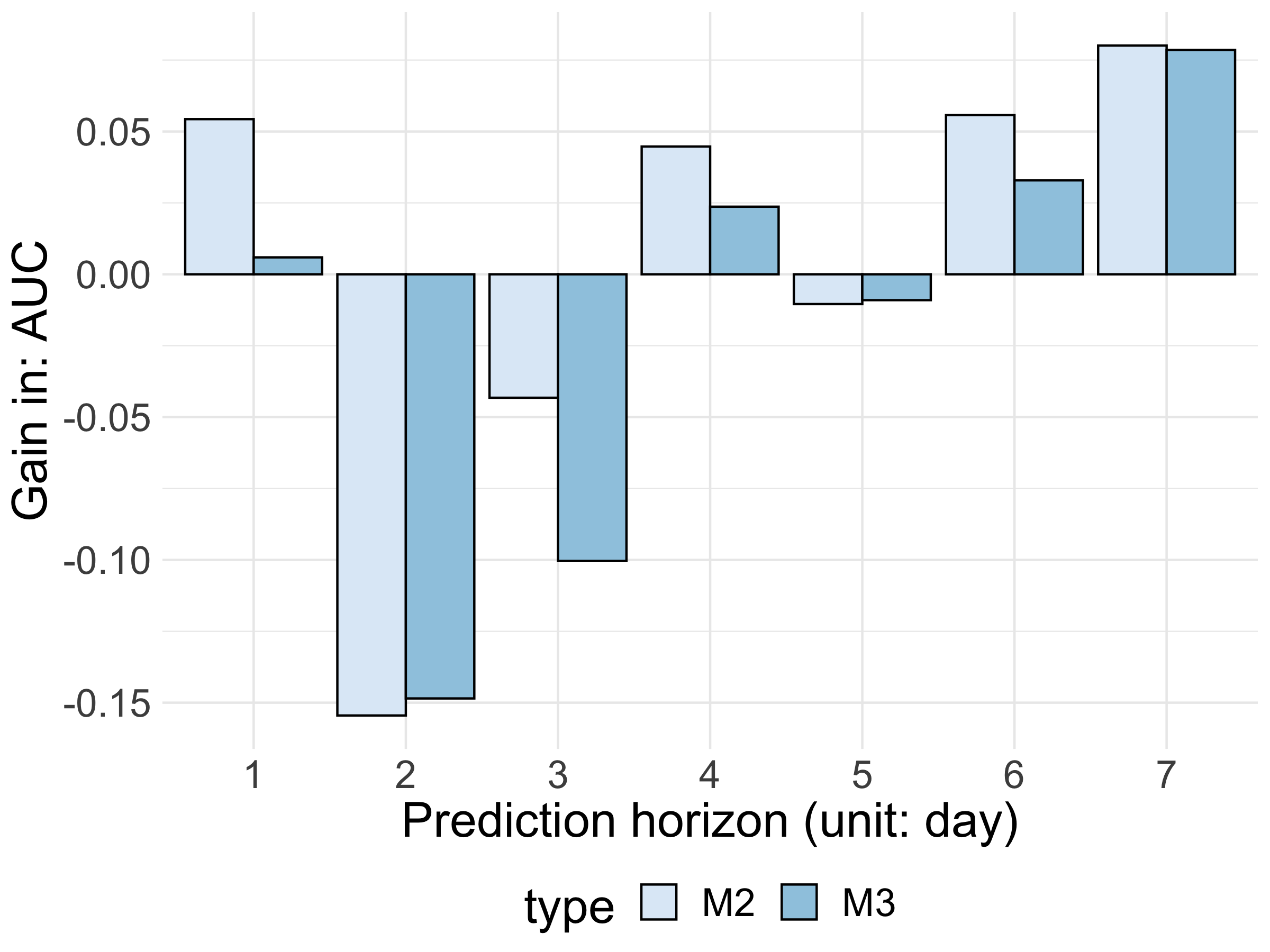}
  \caption{Gains in AUC for PLs with respect to baseline model $M_1$.}
  \label{fig:AUC_pl}
\end{subfigure}~
\begin{subfigure}{.32\textwidth}
  \centering 
  \includegraphics[width=.95\linewidth]{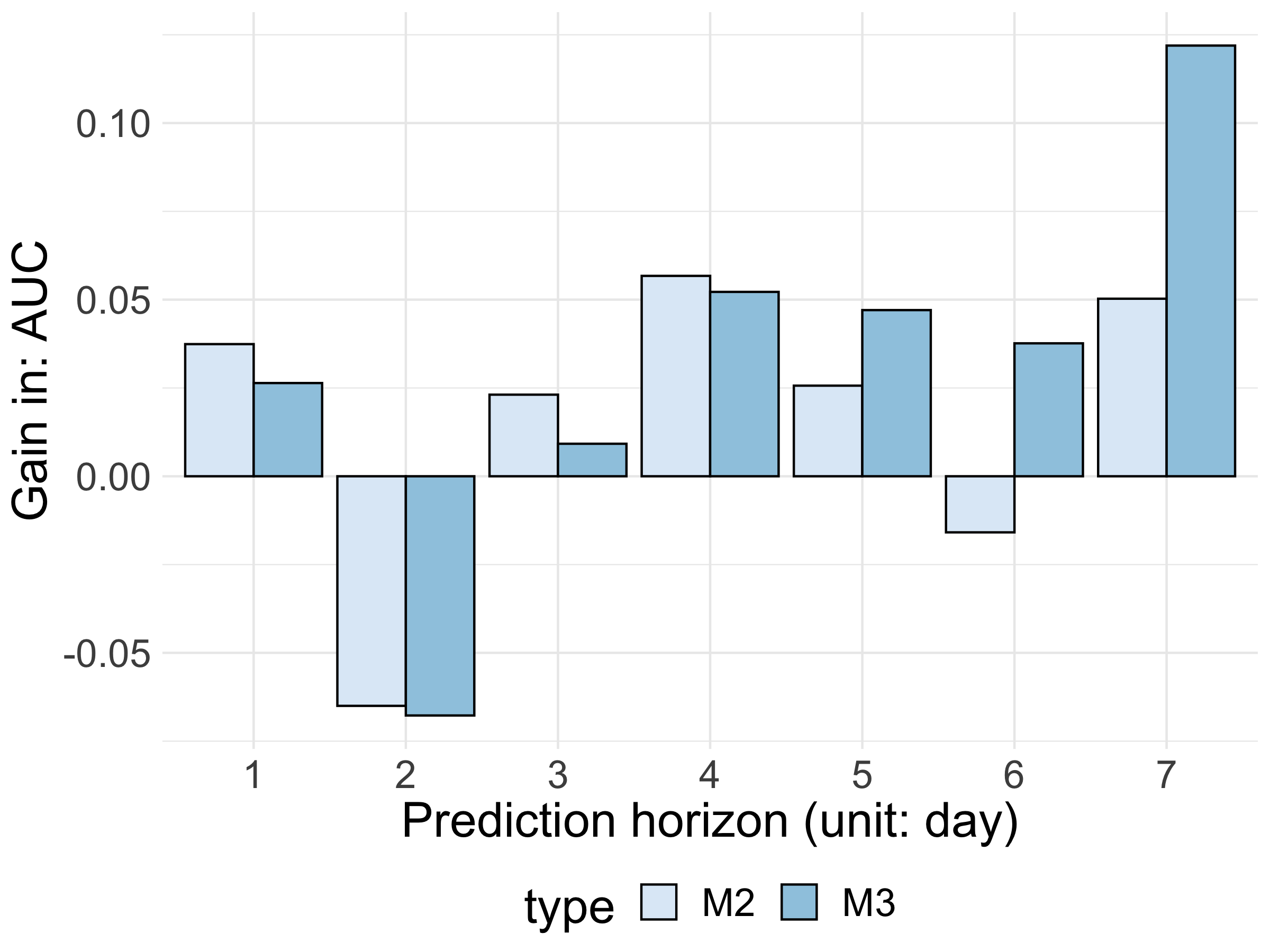}
  \caption{Gains in AUC for PIs with respect to baseline model $M_1$.}
  \label{fig:AUC_pi}
\end{subfigure}
 \vspace{-0.1cm}
\caption{\footnotesize Gains in AUC of the models $M_2$ and $M_3$ containing topological features (based on either VABs or persistence images (PI) or persistence landscapes (PL)) with respect to the baseline model $M_1$ for prediction horizons $h\in \{1,2,\ldots,7\}$ days. The baseline model $M_1$ is free of topological features. $M_2$ involves topological features of dimension one only, whereas $M_3$ is the full model including all the available features.} 
\label{fig:figs}
\vspace{-0.05cm}
\end{figure*}

We compare model performance using the area under the receiver operating characteristic curve (AUC-ROC) measure. The threshold $\delta$ is set to $0.05$ and 150 most dominant nodes are selected when trimming the transaction graphs. We compute the gains achieved by models $M_2$ and $M_3$ with respect to the baseline $M_1$ for prediction horizons $h=1,2,\ldots,7$ (measured in days) using the formula $$\hbox{Gain}(M_i|M_1)=1-\frac{\hbox{AUC}(M_1)}{\hbox{AUC}(M_i)}, \quad i=2,3.$$
Figure \ref{fig:AUC_betti} shows that for horizons $h\geq 3$, the full model $M_3$ has consistently a positive gain and reaches the largest gain of about 22\% for horizon $h=4$. 

For more comparison, we fit additional models replacing only the VAB in our pipeline with two other topological vector summaries of persistence landscapes (PL) and persistence images (PI). The gains achieved by models $M_2$ and $M_3$ for PLs and PIs are depicted in Figures \ref{fig:AUC_pl} and \ref{fig:AUC_pi}. The AUC gains for PIs are similar to VAB but smaller in magnitude, with a maximum of 12\% achieved for horizon $h=7$. For PLs, the gains are inconsistent, reaching a maximum of only about 7.5\%.  In summary, using the  combined $0$  and $1$-dimensional averaged Betti topological summary features leads to the maximum performance gains over the baseline models for larger prediction horizons.    

	
\subsection{Discussion on Key Points in Experimental Results}
In this section, we illustrated the utility of the proposed vector of averaged Bettis (VAB) topological summary  for anomaly detection and prediction in time-varying graphs. Our results showed that topological summaries provide a unique lens to view the data and generally lead to improved performance over prediction models  based on non-topological features. 
 
In the simulation study, low-dimensional VABs were shown to be sensitive enough to all distributional changes behind the graph generating process which led to overall lowest mean absolute error (MAE) of the estimated change points when compared to the traditional graph summaries and common betti number calculation approaches. In the motivating anomalous Etherium price prediction, VABs yielded higher and more consistent gains in AUC against non-topological features and two other competing topological summaries. 

There are limitations in our  study. First, predictive performance of the proposed anomaly detection method in the Etherium price prediction study may change with alternative learning method. Therefore,  besides the random forest models  we also fitted several time series models via generalized additive models (GAM), adding correlation terms to the graph and topological features. However, the gains in the AUC of larger models relative to the baseline  model were very marginal, inconsistent, and not as pronounced as those obtained from the random forest models. Second, computing VAB gets increasingly costly for large graphs. To reduce this computation cost, we  trimmed the Etherieum graphs  by retaining the top most active nodes in order to reduce the sparsity of graphs.   


\section{Conclusion}
\label{sec:conclusion}
In this work, we have proposed a computationally efficient framework to extract topological signatures from a graph with quantitative node attributes. Our framework bypasses the embedding of a graph into a metric to extract topological information.

The extraction of topological information from a graph is done directly and proceeds in two steps: first, we convert the graph into a two-dimensional simplicial complex by adding triangles formed by the graph edges. Then, we compute a persistence diagram (PD) of the lower-star filtration induced by the node attributes. In the second step, we construct the Betti functions (for homological dimensions 0 and 1) associated with the PD and vectorize them over a one-dimensional grid of equally spaced scale values using integration. 

In terms of theoretical contributions, we derive stability results for the proposed vector summary (with respect to the $L_1$ 1-Wasserstein distance) and demonstrate its practical utility in two experimental studies for anomaly detection and prediction in time-varying graphs. 

Though the stability results are proven for a wide class of weights $w$ of a Betti function in (\ref{eqn:betti_function}), in our experiments we set all the weights equal to one. For future research, we plan to experiment with various non-constant weight functions and explore our framework's performance in graph classification and clustering tasks.


\bibliographystyle{unsrt}  
\bibliography{graphtime.bib}

\begin{thebibliography}{10}

\bibitem{victor2020address}
Friedhelm Victor.
\newblock Address clustering heuristics for ethereum.
\newblock In {\em International conference on financial cryptography and data
  security}, pages 617--633. Springer, 2020.

\bibitem{victor2021alphacore}
Friedhelm Victor, Cuneyt~G Akcora, Yulia~R Gel, and Murat Kantarcioglu.
\newblock Alphacore: Data depth based core decomposition.
\newblock In {\em Proceedings of the 27th ACM SIGKDD Conference on Knowledge
  Discovery \& Data Mining}, pages 1625--1633, 2021.

\bibitem{pontiveros2019mint}
Beltran Borja~Fiz Pontiveros, Mathis Steichen, and Radu State.
\newblock Mint centrality: A centrality measure for the bitcoin transaction
  graph.
\newblock In {\em 2019 IEEE International Conference on Blockchain and
  Cryptocurrency (ICBC)}, pages 159--162. IEEE, 2019.

\bibitem{abay2019chainnet}
Nazmiye~Ceren Abay, Cuneyt~Gurcan Akcora, Yulia~R Gel, Murat Kantarcioglu,
  Umar~D Islambekov, Yahui Tian, and Bhavani Thuraisingham.
\newblock Chainnet: Learning on blockchain graphs with topological features.
\newblock In {\em 2019 IEEE international conference on data mining (ICDM)},
  pages 946--951. IEEE, 2019.

\bibitem{li2020dissecting}
Yitao Li, Umar Islambekov, Cuneyt Akcora, Ekaterina Smirnova, Yulia~R Gel, and
  Murat Kantarcioglu.
\newblock Dissecting ethereum blockchain analytics: What we learn from topology
  and geometry of the ethereum graph?
\newblock In {\em Proceedings of the 2020 SIAM international conference on data
  mining}, pages 523--531. SIAM, 2020.

\bibitem{edelsbrunner2010computational}
Herbert Edelsbrunner and John Harer.
\newblock {\em Computational topology: an introduction}.
\newblock American Mathematical Soc., 2010.

\bibitem{carlsson2009topology}
Gunnar Carlsson.
\newblock Topology and data.
\newblock {\em Bulletin of the American Mathematical Society}, 46(2):255--308,
  2009.

\bibitem{zomorodian2005computing}
Afra Zomorodian and Gunnar Carlsson.
\newblock Computing persistent homology.
\newblock {\em Discrete \& Computational Geometry}, 33(2):249--274, 2005.

\bibitem{edelsbrunner2008persistent}
Herbert Edelsbrunner, John Harer, et~al.
\newblock Persistent homology-a survey.
\newblock {\em Contemporary mathematics}, 453:257--282, 2008.

\bibitem{qaiser2019fast}
Talha Qaiser, Yee-Wah Tsang, Daiki Taniyama, Naoya Sakamoto, Kazuaki Nakane,
  David Epstein, and Nasir Rajpoot.
\newblock Fast and accurate tumor segmentation of histology images using
  persistent homology and deep convolutional features.
\newblock {\em Medical image analysis}, 55:1--14, 2019.

\bibitem{smith2021topological}
Alexander~D Smith, Pawe{\l} D{\l}otko, and Victor~M Zavala.
\newblock Topological data analysis: concepts, computation, and applications in
  chemical engineering.
\newblock {\em Computers \& Chemical Engineering}, 146:107202, 2021.

\bibitem{carriere2020topological}
Mathieu Carri{\`e}re and Ra{\'u}l Rabad{\'a}n.
\newblock Topological data analysis of single-cell hi-c contact maps.
\newblock In {\em Topological Data Analysis}, pages 147--162. Springer, 2020.

\bibitem{umeda2017time}
Yuhei Umeda.
\newblock Time series classification via topological data analysis.
\newblock {\em Information and Media Technologies}, 12:228--239, 2017.

\bibitem{pike2020topological}
Jeremy~A Pike, Abdullah~O Khan, Chiara Pallini, Steven~G Thomas, Markus Mund,
  Jonas Ries, Natalie~S Poulter, and Iain~B Styles.
\newblock Topological data analysis quantifies biological nano-structure from
  single molecule localization microscopy.
\newblock {\em Bioinformatics}, 36(5):1614--1621, 2020.

\bibitem{li2019topological}
Max~Z Li, Megan~S Ryerson, and Hamsa Balakrishnan.
\newblock Topological data analysis for aviation applications.
\newblock {\em Transportation Research Part E: Logistics and Transportation
  Review}, 128:149--174, 2019.

\bibitem{cohen2007stability}
David Cohen-Steiner, Herbert Edelsbrunner, and John Harer.
\newblock Stability of persistence diagrams.
\newblock {\em Discrete \& computational geometry}, 37(1):103--120, 2007.

\bibitem{chazal2014persistence}
Fr{\'e}d{\'e}ric Chazal, Vin De~Silva, and Steve Oudot.
\newblock Persistence stability for geometric complexes.
\newblock {\em Geometriae Dedicata}, 173(1):193--214, 2014.

\bibitem{cohen2010lipschitz}
David Cohen-Steiner, Herbert Edelsbrunner, John Harer, and Yuriy Mileyko.
\newblock Lipschitz functions have l p-stable persistence.
\newblock {\em Foundations of computational mathematics}, 10(2):127--139, 2010.

\bibitem{bubenik2015statistical}
Peter Bubenik.
\newblock Statistical topological data analysis using persistence landscapes.
\newblock {\em The Journal of Machine Learning Research}, 16(1):77--102, 2015.

\bibitem{adams2017persistence}
Henry Adams, Tegan Emerson, Michael Kirby, Rachel Neville, Chris Peterson,
  Patrick Shipman, Sofya Chepushtanova, Eric Hanson, Francis Motta, and Lori
  Ziegelmeier.
\newblock Persistence images: A stable vector representation of persistent
  homology.
\newblock {\em The Journal of Machine Learning Research}, 18(1):218--252, 2017.

\bibitem{chazal2021introduction}
Fr{\'e}d{\'e}ric Chazal and Bertrand Michel.
\newblock An introduction to topological data analysis: fundamental and
  practical aspects for data scientists.
\newblock {\em Frontiers in artificial intelligence}, 4, 2021.

\bibitem{hajij2017persistent}
Mustafa Hajij, Bei Wang, Carlos Scheidegger, and Paul Rosen.
\newblock Persistent homology guided exploration of time-varying graphs.
\newblock {\em arXiv preprint arXiv:1707.06683}, 2017.

\bibitem{chung2022persistence}
Yu-Min Chung and Austin Lawson.
\newblock Persistence curves: A canonical framework for summarizing persistence
  diagrams.
\newblock {\em Advances in Computational Mathematics}, 48(1):1--42, 2022.

\bibitem{nanda2014simplicial}
Vidit Nanda and Radmila Sazdanovi{\'c}.
\newblock Simplicial models and topological inference in biological systems.
\newblock In {\em Discrete and topological models in molecular biology}, pages
  109--141. Springer, 2014.

\bibitem{ghrist2008barcodes}
Robert Ghrist.
\newblock Barcodes: the persistent topology of data.
\newblock {\em Bulletin of the American Mathematical Society}, 45(1):61--75,
  2008.

\bibitem{edelsbrunner2022computational}
Herbert Edelsbrunner and John~L Harer.
\newblock {\em Computational topology: an introduction}.
\newblock American Mathematical Society, 2022.

\bibitem{kerber2017geometry}
Michael Kerber, Dmitriy Morozov, and Arnur Nigmetov.
\newblock Geometry helps to compare persistence diagrams, 2017.

\bibitem{kuhn1955hungarian}
Harold~W Kuhn.
\newblock The hungarian method for the assignment problem.
\newblock {\em Naval research logistics quarterly}, 2(1-2):83--97, 1955.

\bibitem{bubenik2018topological}
Peter Bubenik and Tane Vergili.
\newblock Topological spaces of persistence modules and their properties.
\newblock {\em Journal of Applied and Computational Topology}, 2(3):233--269,
  2018.

\bibitem{mileyko2011probability}
Yuriy Mileyko, Sayan Mukherjee, and John Harer.
\newblock Probability measures on the space of persistence diagrams.
\newblock {\em Inverse Problems}, 27(12):124007, 2011.

\bibitem{kraetzl2005random}
Miro Kraetzl, Christine Nickel, and Edward~R Scheinerman.
\newblock Random dot product graphs: a model for social networks.
\newblock {\em Preliminary manuscript}, 2005.

\bibitem{matteson2014nonparametric}
David~S Matteson and Nicholas~A James.
\newblock A nonparametric approach for multiple change point analysis of
  multivariate data.
\newblock {\em Journal of the American Statistical Association},
  109(505):334--345, 2014.

\bibitem{wood2014ethereum}
Gavin Wood.
\newblock Ethereum: A secure decentralised generalised transaction ledger.
\newblock {\em Ethereum project yellow paper}, 151:1--32, 2014.

\bibitem{lopez2009concept}
Sara L{\'o}pez-Pintado and Juan Romo.
\newblock On the concept of depth for functional data.
\newblock {\em Journal of the American statistical Association},
  104(486):718--734, 2009.

\end{thebibliography}

\end{document}